\documentclass{article}

\usepackage{hyperref}
\usepackage{amsmath}
\usepackage{microtype}
\usepackage{graphicx}
\usepackage{subfigure}
\usepackage{booktabs} % for professional tables
\usepackage[utf8]{inputenc}
\usepackage{adjustbox}
\usepackage{cleveref}
\crefname{section}{§}{§§}
\Crefname{section}{§}{§§}

\usepackage{microtype}
\usepackage{graphicx}
\usepackage{subfigure}
\usepackage{booktabs} % for professional tables
\usepackage{colortbl} 
\usepackage{xcolor}
\usepackage{array}

\usepackage[utf8]{inputenc} % allow utf-8 input
\usepackage[T1]{fontenc}    % use 8-bit T1 fonts
\usepackage{hyperref}       % hyperlinks
\usepackage{url}            % simple URL typesetting
\usepackage{booktabs}       % professional-quality tables
\usepackage{amsfonts}       % blackboard math symbols
\usepackage{nicefrac}       % compact symbols for 1/2, etc.
\usepackage{microtype}      % microtypography
\usepackage{xcolor}         % colors
\usepackage{threeparttable}
\usepackage{array}
\usepackage{microtype}
\usepackage{graphicx}
\usepackage{subfigure}
\usepackage{booktabs} % for professional tables
\usepackage{textcomp}
\usepackage{hhline}
\usepackage{color}
\usepackage{multirow}
\usepackage{bbm}
\usepackage{bm}
\usepackage{amsfonts,amssymb}

\usepackage{graphicx} %use graph format
\usepackage{epstopdf}

\usepackage[all]{xy}
\usepackage{amsmath}
\usepackage{wrapfig}
\usepackage{wrapfig,booktabs}

\newcommand{\tabincell}[2]{\begin{tabular}{@{}#1@{}}#2\end{tabular}}

\usepackage{algorithmic}
\usepackage{algorithm}
\usepackage{amsmath}
\usepackage[utf8]{inputenc} % allow utf-8 input
\usepackage[T1]{fontenc}    % use 8-bit T1 fonts
\usepackage{hyperref}       % hyperlinks
\usepackage{url}            % simple URL typesetting
\usepackage{booktabs}       % professional-quality tables
\usepackage{amsfonts}       % blackboard math symbols
\usepackage{nicefrac}       % compact symbols for 1/2, etc.
\usepackage{microtype}      % microtypography
\usepackage{xcolor}         % colors
\usepackage{stfloats}
\usepackage{float}  
\usepackage{wrapfig}
\usepackage{wrapfig,lipsum,booktabs}
\usepackage{amsmath}

\newcommand{\cv}[0]{\ensuremath{\boldsymbol{c}} }
\newcommand{\dv}[0]{\ensuremath{\boldsymbol{d}} }

\newcommand{\fv}[0]{\ensuremath{\boldsymbol{f}} }
\newcommand{\gv}[0]{\ensuremath{\boldsymbol{g}} }
\newcommand{\hv}[0]{\ensuremath{\boldsymbol{h}} }
\newcommand{\iv}[0]{\ensuremath{\boldsymbol{i}} }

\newcommand{\kv}[0]{\ensuremath{\boldsymbol{k}} }

\newcommand{\pv}[0]{\ensuremath{\boldsymbol{p}} }

\newcommand{\rv}[0]{\ensuremath{\boldsymbol{r}} }

\newcommand{\xv}[0]{\ensuremath{\boldsymbol{x}} }

\newcommand{\zv}[0]{\ensuremath{\boldsymbol{z}} }

\newcommand{\Iv}[0]{\ensuremath{\boldsymbol{I}} }

\newcommand{\Wv}[0]{\ensuremath{\boldsymbol{W}} }

\newcommand{\Phiv}[0]{\ensuremath{\boldsymbol{\Phi}} }
\newcommand{\betav}[0]{\ensuremath{\boldsymbol{\beta}} }
\newcommand{\thetav}[0]{\ensuremath{\boldsymbol{\theta}} }
\newcommand{\lambdav}[0]{\ensuremath{\boldsymbol{\lambda}} }
\newcommand{\epsilonv}[0]{\ensuremath{\boldsymbol{\epsilon}} }

\newcommand{\cdotv}[0]{\ensuremath{\boldsymbol{\cdot}}}

\newcommand{\given}{\,|\,}

\usepackage[preprint]{neurips_2024}

\usepackage[utf8]{inputenc} % allow utf-8 input
\usepackage[T1]{fontenc}    % use 8-bit T1 fonts       % hyperlinks
\usepackage{url}            % simple URL typesetting
\usepackage{booktabs}       % professional-quality tables
\usepackage{amsfonts}       % blackboard math symbols
\usepackage{nicefrac}       % compact symbols for 1/2, etc.
\usepackage{microtype}      % microtypography
\usepackage{xcolor}         % colors

\title{A Non-negative VAE: \\ the Generalized Gamma Belief Network}

\author{\hspace{1mm}Zhibin Duan \\
Xidian University\\
\texttt{xd\_zhibin@163.com} \\
	\And
\hspace{1mm}Tiansheng Wen \\
Xidian University\\
\texttt{neilwen987@stu.xidian.edu.cn} \\
	\And
 \hspace{1mm}Muyao Wang \\
Xidian University\\
\texttt{muyaowang@stu.xidian.edu.cn} \\
\And
\hspace{1mm}Bo Chen\thanks{Corresponding author} \\
Xidian University\\
\texttt{bchen@mail.xidian.edu.com} \\
\And
\hspace{1mm}Mingyuan Zhou \\
The University of Texas at Austin\\
\texttt{mingyuan.zhou@mccombs.utexas.edu} \\
}
\begin{document}

\maketitle

% The gamma belief network (GBN), often seen as a deep topic model, has shown its potential for exploring text multi-layer interpretable latent representation.
% Its outstanding ability to learn interpretable latent factors can be partly 
% attributed to sparse and non-negative gamma-distributed latent variables.
% However, the existing GBN and its variants are equipped with the linear generative model, which limits their expressiveness and applications.
% To remove this limitation, we develop the generalized gamma belief network (Generalised GBN) in this paper, which generalizes the original linear generative model to the more expressive non-linear generative model.
% Since the parameters of the Generalized GBN no longer have an analytic conditional posterior, we further build an upward-downward Weibull inference network to approximate the latent variable's posterior.
% The generative model and inference network parameters are jointly trained under the variational inference framework.
% In the end, we conduct extensive experiments on both expressivity and disentangled representation learning tasks to verify the performance of 
% generalized GBN with state-of-the-art Gaussian variational autoencoders as the baselines.
\begin{abstract}
The gamma belief network (GBN), often regarded as a deep topic model, has demonstrated its potential for uncovering multi-layer interpretable latent representations in text data. Its notable capability to acquire interpretable latent factors is partially attributed to sparse and non-negative gamma-distributed latent variables. However, the existing GBN and its variations are constrained by the linear generative model, thereby limiting their expressiveness and applicability. To address this limitation, we introduce the generalized gamma belief network (Generalized GBN) in this paper, which extends the original linear generative model to a more expressive non-linear generative model. Since the parameters of the Generalized GBN no longer possess an analytic conditional posterior, we further propose an upward-downward Weibull inference network to approximate the posterior distribution of the latent variables. The parameters of both the generative model and the inference network are jointly trained within the variational inference framework. Finally, we conduct comprehensive experiments on both expressivity and disentangled representation learning tasks to evaluate the performance of the Generalized GBN against state-of-the-art Gaussian variational autoencoders serving as baselines.
\end{abstract}
\section{Introduction}
\label{submission}
Variational autoencoders (VAEs) \citep{kingma2013auto, rezende2014stochastic}, which marry the expressiveness of deep neural networks with the robustness of stochastic latent variables, have gained great success in the last ten years.
As a class of probabilistic generative models, VAEs are popularly used in generation tasks ranging from image generation \citep{vahdat2020nvae} to text generation \citep{bowman2015generating} to graph generation \citep{kipf2016variational}.
In addition to their productive generative abilities, VAEs enjoy favorable properties in extracting meaningful and interpretable factorized representation, leading to their utilization in representation learning \citep{bengio2013representation,  higgins2016beta, lake2017building,srivastava2017autoencoding}.
% Another attractive property of VAE is that it can learn interpretable factorized representation in an unsupervised manner, which is critical for building human-like AI systems \citep{bengio2013representation, higgns2016beta, lake2017building}.
Benefiting from VAE's attractive characteristics, many efforts have been made to improve its expressivity  \citep{van2016conditional, vahdat2020nvae, child2020very} and disentangled representation learning capability further \citep{higgins2016beta, kim2018disentangling,chen2018isolating, kumar2017variational}.
% With long-range efforts, the performance in the likelihood of VAEs can outperform state-of-the-art PixelCNN-based autoregressive models \citep{van2016conditional, child2020very}.
% Besides, the works for learning disentangled representation have also attracted wide interest and have made significant advancements \citep{higgins2016beta, kim2018disentangling,chen2018isolating, kumar2017variational}.
% Besides, the works for learning disentangled representation have also attracted wide interest 
% and achieved great progress \citep{higgins2016beta, kim2018disentangling,chen2018isolating, kumar2017variational}. 
% to employing attention mechanisms \citep{apostolopoulou2021deep} 

% To be more precise, adding more stochastic layers is a straightforward method to improve VAE's generative capacity \citep{sonderby2016ladder}. However, it is not direct due to the well-known ``latent variable collapse'' issue, in which the approximate posterior of the higher layer is independent of the data \citep{dieng2019avoiding}. 
% To build effective hierarchical VAEs, there is a lot of effort, ranging from building a novel deep generative model \citep{sonderby2016ladder, maaloe2019biva} to reconstructing network structure \citep{vahdat2020nvae, child2020very} to developing more expressive variational distributions \citep{vahdat2020nvae}.

Parallel to the development of Gaussian VAEs, adequate progress has been achieved on the gamma belief network (GBN) \citep{zhou2015poisson, zhou2016augmentable}.
In particular, the GBN, as a deep Bayesian factor analysis model, has the appealing property of learning interpretable multi-level latent representations from concrete to abstract \citep{lee1999learning}.
% Benefiting from the gamma distribution's ability to model sparse, the GBNs have greatly succeeded in modeling sparse count data, especially the text bag-of-words representation \citep{zhou2015poisson,panda2019deep}.
% factorize a count matrix under the Poisson likelihood and further factorize the
% shape parameters of the gamma-distributed hidden units of
% each layer into the product of a connection weight matrix
% and the gamma-hidden units of the next layer.
%  in
%  readily interpretable multilayer
% latent representations.
To marry the expressiveness of the deep neural network, \cite{zhang2020deep} extends GBN by utilizing the Weibull variational inference network to approximate the posterior of its gamma latent variables, resulting in a deep variational autoencoder.
For modeling document bag-of-words representation, the GBN and its variants have achieved attractive performance in generative ability and extracting interpretable latent factors.
The outstanding performance can be attributed to the sufficient ability of the gamma distribution to model sparsity \citep{zhang2018whai}.
Furthermore, recent studies have been performed to enhance the expressiveness of GBN by incorporating deeper stochastic layers, resulting in impressive progress \citep{duan2021sawtooth, li2022alleviating, duan2023bayesian}.

\begin{figure} [t!]
	\centering
    \subfigure[Linear Layer]{\includegraphics[height = 3.4cm,width = 3.4cm]{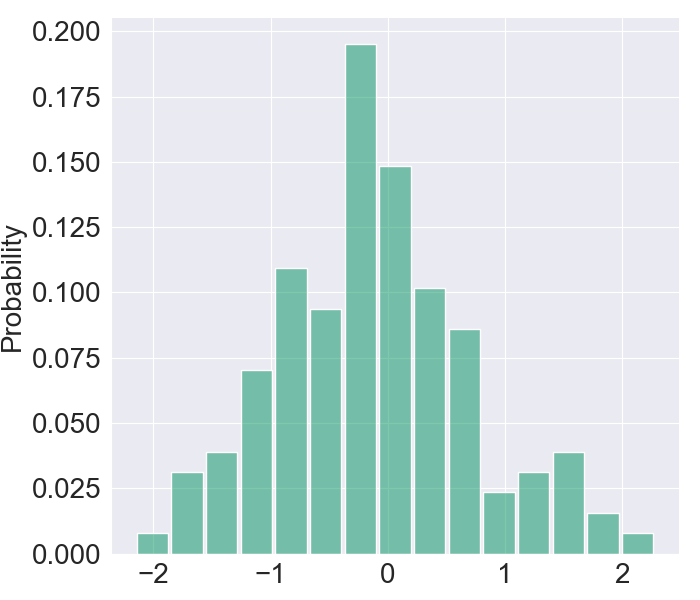}\label{motivation_1}}
    \subfigure[Gaussian] { \includegraphics[height = 3.4cm,width = 3.4cm]{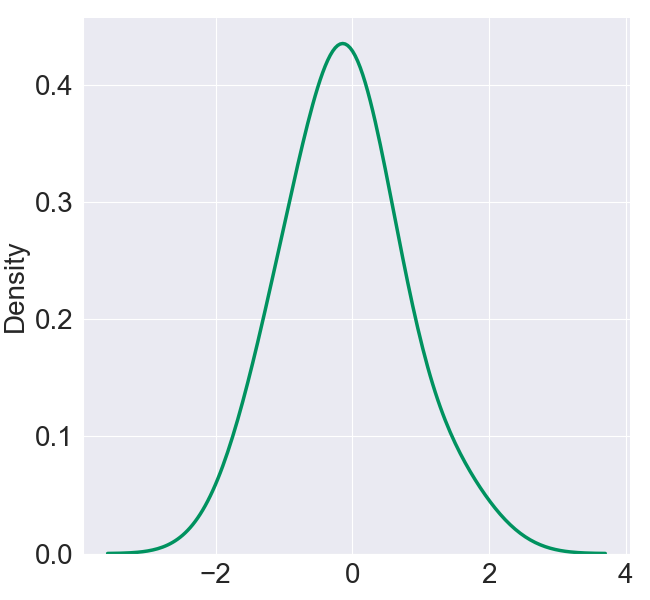}\label{motivation_3}}
    \subfigure[Activation Layer]{\includegraphics[height = 3.4cm,width = 3.4cm]{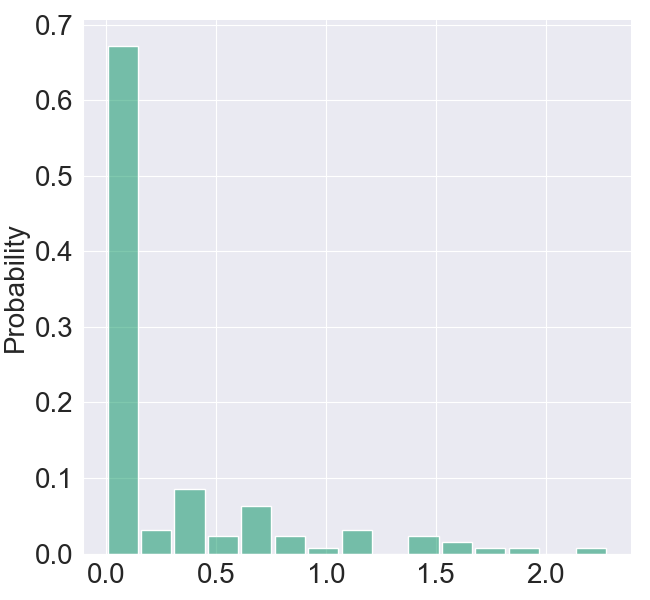}\label{motivation_2}} 
    \subfigure[Gamma]{\includegraphics[height = 3.4cm,width = 3.4cm]{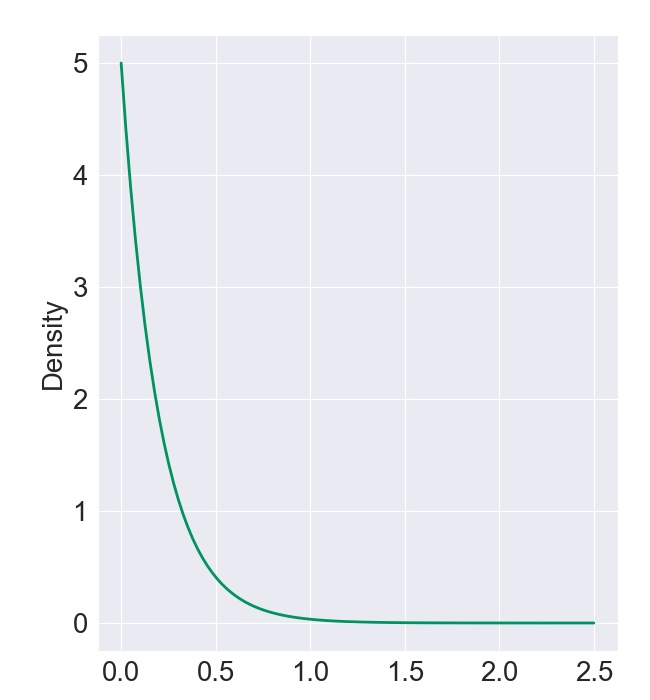}\label{motivation_4}}
    \caption{
    With a trained autoencoder on the MNIST dataset, the probability histogram for the hidden representation is displayed in \textit{\ref{motivation_1}} as the output from a linear layer and \textit{\ref{motivation_2}} as the output from a ReLU activation layer.
    The former is suitable to be approximated by Gaussian distribution 
    (\textit{\ref{motivation_3}}), while the latter is suitable to be approximated by Gamma distribution (\textit{\ref{motivation_4}}).   } 
    \label{motivation}
    \vspace{-5mm}
\end{figure}

While GBN and its variants have achieved great progress, they are limited to the linear generative model, restricting their expressiveness and applications \citep{zhou2015poisson, zhang2018whai, wang2020deep, wang2022generative}.
The counterpart is the Gaussian VAE, which can employ more expressive neural networks as generative models (decoders) to improve its expressivity \citep{vahdat2020nvae, child2020very}.
From another perspective, benefiting from the sparse and non-negative latent variables, the gamma latent model has potential advantages in learning disentangled representation \citep{lee1999learning, bengio2013representation, tonolini2020variational, mathieu2019disentangling}.
While a series of works have been developed to improve Gaussian VAE's disentangled representation learning ability \citep{higgins2016beta, chen2018isolating}, they often rely on adding different regularizers, which may hurt the test reconstruction performance \citep{kim2018disentangling, mathieu2019disentangling}.
Moreover, introducing non-negative and sparse gamma latent variables to neural networks is natural and reasonable.
Generally, most neural networks are composed of linear feed-forward units and non-linear activation units, in which the former outputs dense real vectors and the latter outputs sparse non-negative vectors \citep{nair2010rectified}, as illustrated in Fig.~\ref{motivation_1} and Fig.~\ref{motivation_2}.
While the Gaussian distribution is suitable for modeling dense real vectors, the gamma distribution is suitable for modeling sparse non-negative vectors, as demonstrated in Fig.~\ref{motivation_3} and Fig.~\ref{motivation_4}.

% \citep{geva2020transformer, dai2021knowledge, shen2023study} points out that the non-negative sparse vector after activation function can be seen as an un-normalized attention weight, which can be used to identify the neurons that express the fact.
All the considerations in the last section motivate us to explore gamma latent variable models further.
Specifically, to remove the GBN's limitation on the linear generative model, we develop the generalized gamma belief network (Generalized GBN), which employs a non-linear network to improve the model's expressiveness and retain interpretation with sparse and non-negative gamma latent variables.
% Compared with existing GBNs, the generalized GBN mainly removes the limitation of linear decoding, and compared with exciting Gaussian VAEs, the generalized GBN replaces the gamma distribution with the Gaussian distribution in the generative model.
Due to the fact that the gamma distribution can't be reparameterized,  we follow the former works \citep{zhang2018whai} that use the Weibull distribution to approximate its posterior in the generative model.
To verify the effectiveness of the proposed model, we conduct two kinds of experiments to evaluate its expressivity and disentangled latent representation learning capabilities, respectively.
The results of the expressivity experiment indicate that the Generalized GBN can perform on par with state-of-the-art hierarchical Gaussian VAEs on test likelihood.
The disentangled experiment results verify our motivation that using spare and non-negative gamma latent variables can effectively enhance the generative model's disentangled ability.
The main contributions of the paper can be summarized as follows:
\vspace{-3pt}
\begin{itemize}
\setlength{\itemsep}{3pt}
\setlength{\parsep}{0pt}
\setlength{\parskip}{0pt}
\item With the intention of removing the linear generative model constraint of the GBN, we construct the generalized gamma belief network, which can be equipped with more expressive non-linear neural networks as the generative model (decoder).
\item  To approximate the posterior of latent variables in the Generalized GBN, we design the Weibull variational upward-downward inference network.
\item To verify the expressivity and disentangled representation learning capabilities of the Generalized GBN, we conducted extensive experiments on the benchmark datasets. 
\end{itemize}

\begin{figure*}[!ht]
\centering
\subfigure[GBN]{
\includegraphics[width=0.26\linewidth]{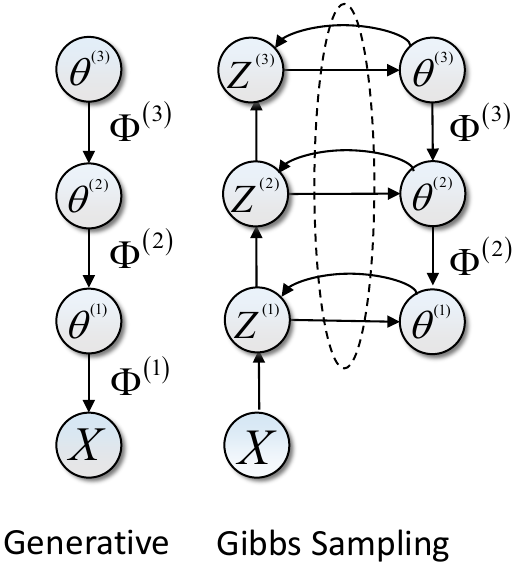}
\label{fig_gbn}
}
\quad \quad  \quad  \quad 
\subfigure[Hierarchical VAE]{
\includegraphics[width=0.21\linewidth]{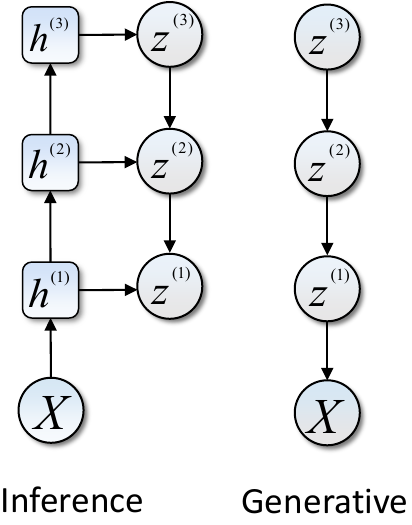}
\label{fig_hvae}
} 
\quad \quad   \quad  \quad 
\subfigure[Generalized GBN]{
\includegraphics[width=0.205\linewidth]{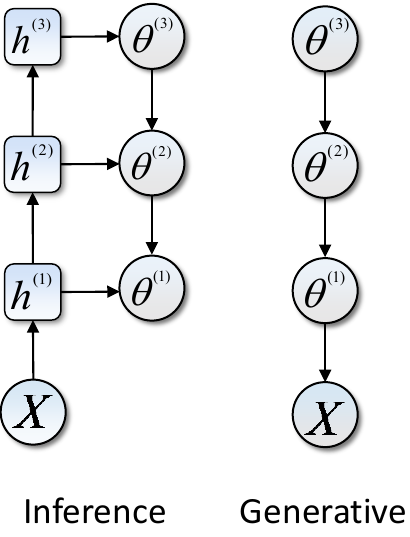}
\label{fig_ggbn}
} \vspace{-3mm}
\caption{The graphical model of \textit{~\ref{fig_gbn}:} the generative model of Gamma Belief Network (GBN), and a sketch of the upward-downward Gibbs sampler, where $Z^{(l)}$ are augmented latent counts that are upward sampled in each Gibbs sampling
iteration; \textit{\ref{fig_hvae}:} the generative model and inference network of the hierarchical Gaussian VAE;  \textit{\ref{fig_ggbn}:}  the inference and generation of the Generalized gamma belief network (Generalized GBN).
Circles are stochastic variables, and squares are deterministic variables. 
% The orange and
% blue arrows denote the upward and downward information propagation respectively, and the red
% ones denote the data generation.
 } \label{fig_inference_net}
\vspace{-4mm}
\end{figure*} 
\vspace{-2mm}
\section{Preliminaries}
\vspace{-2mm}
This section will give a detailed description of the gamma belief network and Gaussian VAEs.
\vspace{-1mm}
\subsection{Gamma Belief Network}
Assuming the observations are multivariate count vectors ${\small x_j \in {\mathbb{Z}^{{K_0}}}}$, as shown in Fig.\ref{fig_gbn}, the generative model of the gamma belief network (GBN) \citep{zhou2015poisson, zhou2016augmentable} with $T$ hidden layers, from top to bottom, is expressed as
\begin{equation}
\scalebox{0.95}{$ 
\begin{split}
\label{eq_gbn}
&\bm{\theta}_j^{(L)} \sim \mbox{Gamma}\left(\rv, \, \cv_j^{(L + 1)}\right),  \cdots, \bm{\theta}_j^{(l)} \sim \mbox{Gamma}\left(\bm{\Phi}^{(l + 1)} \bm{\theta}_j^{(l + 1)}, \, \cv_j^{(l + 1)}\right), \cdots,\\[3pt]
&\bm{\theta}_j^{(1)} \sim \mbox{Gamma}\left(\bm{\Phi}^{(2)} \bm{\theta}_j^{(2)}, \, \cv_j^{(2)}\right), \bm{x} _j \sim \mbox{Poisson}\left(\bm{\Phi} ^{(1)} \bm{\theta} _j ^{(1)}\right), 
\end{split}$}
\end{equation}where, the count vector $\xv_j$ (e.g., the bag-of-word of document $j$) is factorized as the product of the factor loading matrix ${\small \bm{\Phi}^{(1)} \in \mathbb{R}^{K_l}_+ }$ (topics), and gamma distributed factor scores ${\small \bm{\theta} _j^{(1)}\in \mathbb{R}^{K_1}_+}$ (topic proportions), under the Poisson likelihood; and the hidden units ${\small \bm{\theta} _j^{(l)} \in \mathbb{R}^{K_l}_+}$ of layer $l$ is further factorized into the product of the factor loading ${\small \bm{\Phi} ^{(l+1)} \in \mathbb{R}_+^{K_{l} \times K_{l+1}}}$ and hidden units of the next layer to infer a multi-layer latent representation; 
the top layer's hidden units share the same vector as their gamma-shape parameters.
and the $p_j^{(2)}$ are probability parameters and ${\small \{ 1/c^{(t)}\}_{3, T+1}}$ are gamma scale parameters, with ${\small c_j^{(2)}:= (1-p_j^{(2)})/p_j^{(2)}}$.
% Compared with VAEs, that
% perform approximate inference due to the non-conjugacy of the model,
For scale identifiabilty and ease of inference, each column of ${\small \bm{\Phi} ^{(l)} \in \mathbb{R}_+^{K_{l-1} \times K_{l}}}$ is restricted to have
a unit ${\small L_1}$ norm.
Benefiting from analytic
conditional posteriors for all parameters, GBN, can be inferred via a Gibbs sampler.
\vspace{-1mm}
\subsection{Hierarchical Gaussian VAE}
As shown in Fig.~\ref{fig_hvae},
the generative model of  hierarchical VAE \citep{sonderby2016ladder} for data $\xv_j$ with $L$ layers of Gaussian distribution latent variables $\{\zv_j^{(l)}\}_{l=1}^{L}$, can be described as:
\begin{equation}
\scalebox{0.95}{$
\begin{split}
&\zv_{j}^{\left( L \right)}\sim \mathcal{N} \left( {\bf{0},\, \Iv} \right), \cdots ,{\zv_{j}^{\left( l \right)}}\sim \mathcal{N} \left( {\gv_{\mu }^{\left( {l + 1} \right)}( {{\zv^{( {l + 1} )}_j}}) , \, \gv_{\sigma }^{\left( {i + 1} \right)}({{\zv^{\left( {l + 1} \right)}_j}})} \right), \cdots ,\\[3pt]
&{\zv_{j}^{\left( 1 \right)}}\sim \mathcal{N} \left( {\gv_{\mu }^{\left( {2} \right)}( {{\zv^{( {2} )}_j}}) , \, \gv_{\sigma }^{\left( {2} \right)}({{\zv^{\left( {2} \right)}_j}})} \right), \xv_{j} \sim \mathcal{N}\left( {\gv_{\mu }^{\left( 1 \right)}( {{\zv_{j}^{\left( 1 \right)}}} ),  \, \gv_{\sigma }^{\left( 1 \right)}( {{\zv_{j}^{\left( 1 \right)}}} )} \right),
\end{split}$}
\end{equation}
where the observation model is matching continuous-valued data, the ${\small \gv_{\cdotv}^{(l)}(\cdotv)}$ are the non-linear neural networks in generative models.
Since the parameters in VAEs don't have analytic conditional posteriors, they need to be inferred by variational inference \citep{kingma2013auto}. 
Specifically, conditioned on the stochastic layer below each stochastic layer, which is specified as a fully factorized Gaussian distribution, the variational posterior can be defined as:
\begin{equation}
\scalebox{0.95}{$
\begin{split}
q{(\zv^{\left( l \right)}_j)} =  \mathcal{N} \left( {\fv_{\mu }^{\left( {l + 1} \right)}( {{\xv_j, \zv^{( {l + 1} )}_j}}) , \, \fv_{\sigma}^{\left( {l + 1} \right)}({{\xv_j, \zv^{\left( {l + 1} \right)}_j}})} \right)
\end{split}$}
\end{equation}where the ${\small \fv_{\cdotv}^{(l)}(\cdotv)}$ are non-linear neural networks in variational inference networks.
We can write the variational lower bound
$\mathcal{L}_{\mbox{vae}}(\xv)$ on $\mbox{log} \: p(\xv)$ as 

\begin{equation}
\scalebox{0.95}{$
\begin{split}
\mathcal{L}_{\mbox{vae}}(\mathbf{x}) := & \mathbb{E}_{q(\mathbf{z} \mid \mathbf{x})} \left[ \log p(\mathbf{x} \mid \mathbf{z}^{(1)}) \right] - \sum_{l=1}^L \mathbb{E}_{q(\mathbf{z}^{>l} \mid \mathbf{x})} \left[ \text{KL} \left( q(\mathbf{z}^{(l)} \mid \mathbf{x}, \mathbf{z}>l) \, \| \, p(\mathbf{z}^{(l)} \mid \mathbf{z}^{>l}) \right) \right]
\end{split}$}
\end{equation}
where ${ q\left( {{\zv^{\left( { > l} \right)}}|\,\xv} \right): = \prod\nolimits_{l = 1}^{L - 1} {q\left( {{\zv^{\left( l \right)}}|\,\xv,{\zv^{\left( { > l} \right)}}} \right)}}$ is the approximate posterior up to the $(l-1)^{th}$ layer.
The objective
is trained using the reparameterization trick with the repreameter trick \citep{kingma2013auto, rezende2014stochastic}.
\vspace{-3mm}
\section{The Generalized Gamma Belief Network}
\vspace{-2mm}
This section provides a detailed description of the proposed Generalized GBN, which consists of the hierarchical non-linear generative model (Sec.\ref{subsec: generative_model}) and the variational inference network (Sec.\ref{subsec: inference_network}).
Followed by the description of the variational inference (Sec.\ref{subsec: elbo}) and some techniques for stable training ( Sec.~\ref{subsec_stable_traing}). 
\vspace{-1mm}
\subsection{Hierarchical Generative Model}
\vspace{-1mm}\label{subsec: generative_model}
To generalize GBN with more expressive non-linear generative models (decoders), we take inspiration from the hierarchical Gaussian VAE \citep{vahdat2020nvae, child2020very} to build the  Generalized GBN.
As shown in Fig.~\ref{fig_ggbn},
the generative model of the Generalized GBN with $L$ layers, from top to bottom, can be expressed as
\begin{equation}
\setlength{\abovedisplayskip}{8pt}
\setlength{\belowdisplayskip}{8pt}
\scalebox{0.95}{$
\begin{split}
\label{SawETM}
&\thetav_j^{(L)} \sim \mbox{Gamma}\left(\rv, \cv^{(L + 1)}\right),  \cdots, \thetav_j^{(l)} \sim \mbox{Gamma}\left(\gv_{\alpha }^{\left( {l + 1} \right)} (\thetav_j^{(l + 1)}),\: \gv_{\beta}^{(l+1)}(\thetav_j^{(l+1)})\right),  \cdots, \\[3pt] 
&\thetav_j^{(1)} \sim \mbox{Gamma}\left(\gv_{\alpha }^{\left( {2} \right)} (\thetav_j^{(2)}), \: \gv_{\beta}^{(2)}(\thetav_j^{(2)})\right),  \xv _j \sim \mathcal{N} \left(\gv_{\mu}^{\left( {1} \right)} (\thetav_j^{(1)}), \gv_{\sigma}^{\left( {1} \right)} {(\thetav_j^{(1)})}\right),
\end{split}$}
\end{equation}
where the observation model is matching continuous-valued
data, the $\gv_{\cdotv}^{(l)}(\cdot)$ are non-linear neural networks, such as ResNet \citep{he2016deep}, as decoders.
To satisfy different types of observation data, such as count data and binary data, the Generalized GBN can be adapted directly with different output layers as follows:
\begin{equation}
\scalebox{0.95}{$
\begin{split}
&\xv _j \sim \mbox{Poisson}\left(\gv^{\left( {1} \right)} (\thetav_j^{(1)})\right),  \xv _j \sim \mbox{
Bernoulli}\left(\gv^{\left( {1} \right)} (\thetav_j^{(1)})\right).
\end{split}$}
\end{equation}
% Unlike the Gaussian VAE, the neural network $\gv_{\cdotv}^{(l)}(\cdot)$ in Generalized GBN should output non-negative vectors to satisfy the demand in the Gamma distribution.
To meet the demand in the Gamma distribution, the neural network $\gv_{\cdotv}^{(l)}(\cdot)$ in the Generalized GBN should output non-negative vectors.
Considering the property of continuous differentiability, we employ $\mbox{Softplus}(\cdotv)$ non-linear function to each element to ensure positiveness in our generative model, where $\mbox{Softplus}(\cdotv)
= \mbox{log}(1+\mbox{exp}(\cdot))$.

\textbf{Connection with GBN:}
It's evident that the Generalized GBN will reduce to the GBN \citep{zhou2015poisson}  when ${\gv_{\alpha }^{( {l + 1} )} (\thetav_j^{(l + 1)}) = \Phiv^{(l + 1)}\thetav_j^{(l + 1)}}$
as the non-linear generative model reduces to linear generative models.
\vspace{-1mm}
\subsection{Upward-Downward Variational  Inference Network}
\vspace{-1mm}
\label{subsec: inference_network}
Similar to Gaussian VAE, the parameters of the Generalized GBN do not have analytic conditional posteriors, which require a variational inference network to approximate the latent variable's posterior.

% \textbf{Weibull Approximate Posterior:} 
\textbf{Weibull Approximate Posterior:}
Although the gamma distribution seems logical for the posterior distribution because it encourages sparsity and satisfies the nonnegative condition, it is not reparameterizable and cannot be optimized using gradient descent.
And considering $\iv)$, the Weibull distribution has a simple
reparameterization so that it is easier to optimize, and  $\iv\iv)$ the Weibull distribution is
similar to a gamma distribution, capable of modeling sparse,
skewed and positive distributions, we use the Weibull distribution \citep{zhang2018whai} to approximate the posterior for the gamma latent variables.
Specifically, the latent variable $x\sim \mbox{Weibull}(k,\lambda )$ can be easily reparameterized as:
\begin{equation}
x = \lambda {( - \ln (1 - \varepsilon ))^{1/k}},~~\varepsilon \sim \mbox{Uniform}(0,1).
\label{eq_reparameterize}
\end{equation}
$ii)$, The KL divergence from the gamma to Weibull distributions has an analytic expression as:
\begin{equation} 
\setlength{\abovedisplayskip}{3pt}
\setlength{\belowdisplayskip}{3pt}
\begin{aligned}
\mbox{KL}\left( \mbox{Weibull}(k,\lambda) \,\|\, \mbox{Gamma}(\alpha,\beta) \right) &= \\
 \frac{\gamma \alpha}{k} - \alpha \log \lambda &+ \log k + \beta \lambda \Gamma\left(1+\frac{1}{k}\right) - \gamma - 1 - \alpha \log \beta + \log \Gamma(\alpha).
\end{aligned}
\end{equation} 
where $\gamma$ is the Euler-Mascheroni constant.
% due to its simple reparameterization, and an analytic expression of the Kullback--Leibler (KL) divergence. 
% \textbf{Upward-Downward Inference Network:} 

\textbf{Upward-Downward Inference Network:}
As shown in Fig.~\ref{fig_ggbn}, the variational inference network combines the obtained latent features with the prior from the stochastic up-down path to 
construct the variational posterior:
\begin{equation}
\label{Eq_inference}
\scalebox{0.95}{$
\begin{split}
    q\left(\bm{\theta}_j^{(l)} \given \bm{h}_j^{(l)}, \bm{\theta}_j^{(l+1)}\right) &= \mbox{Weibull}\left(\bm{k}_j^{(l)}, \bm{\lambda}_j^{(l)}\right), \\[3pt]
    \bm{k}_j^{(l)} = \mbox{Softplus}\left(\fv_{k}^{(l)}(\bm{\theta}_j^{(l+1)},{\bm{h}}_j^{(l)})\right),&   \bm{\lambda}_j^{(l)} = \mbox{Softplus}\left(\fv_{\lambda}^{(l)}(\bm{\theta}_j^{(l+1)}, {\bm{h}}_j^{(l)})\right) ,
\end{split}$}
\end{equation}
where $\fv_{\cdotv}^{(l)}(\cdotv)$ denotes the neural network, and $\mbox{Softplus}$ applies $\mbox{log}(1+\mbox{exp}(\cdot))$ nonlinearity to each element to ensure positive Weibull shape and scale parameters. The Weibull distribution is used to approximate the gamma-distributed conditional posterior, and its parameters $\bm{k}_j^{(l)} \in \mbox{R}_{+}^{K_l}$ and $\bm{\lambda}_j^{(l)} \in \mbox{R}_{+}^{K_l}$ are both inferred by combining the bottom-up likelihood information and the prior information from the generative distribution using the neural networks. 
% The inference network is structured as
% %\begin{small}
% \begin{align}\label{inference network}
% q(\bm{\theta} \given \xv)=q(\bm{\theta}^{(L)} \given \xv) \prod_{l=1}^{L-1} q(\bm{\theta}^{(l)} \given \bm{\theta}^{(l+1)},\xv).
% \end{align}
The use of both top-down prior information and bottom-up data information is one of the linkages between the variational inference network and the Gibbs Sampler of GBN, as illustrated in Fig.~\ref{fig_inference_net}.%\end{small}
% compared with Eq.~\eqref{q_theta}, the residual upward pass in SawETM allows all the  latent variables to have a deterministic dependency on input $x$, thus the top stochastic latent variables could receive efficient information and will be empirically less likely to collapse.
\vspace{-1mm}
\subsection{Variational Inference}\vspace{-1mm}\label{subsec: elbo}
For the Generalized GBN, given the model parameters referred to as ${\small \Wv^{(l:L)}}$, which consist of the parameters in the generative model and inference network, the marginal likelihood of the dataset $X$ is defined as:
\begin{equation}
\scalebox{0.95}{$
\begin{split}
&\pv \left( { {X} \given \{ {{\Wv^{\left( l \right)}}} \}_{l = 1}^L} \right)  = \int {\prod\limits_{l = 1}^L { {\prod\limits_{j = 1}^J {\pv\left( {\xv_j \: | \: \thetav_j^{(1)}}\right)} }}}  {\prod\limits_{l = 1}^{L} { {\prod\limits_{j = 1}^J {\pv\left( {\thetav _j^{\left( l \right)} \: | \: \thetav _j^{\left( {l + 1} \right)}} \right)} }}} \dv \thetav _{l = 1,j = 1}^{L,J} .
\end{split}$}
\end{equation}
The inference task is
to learn the parameters of the generative model and the inference network.
Similar to VAEs,
the optimization objective of the Generalized GBN can be achieved by
maximizing the evidence lower bound (ELBO) of log-likelihood as
\begin{equation} \label{aug_ELBO}
\scalebox{0.95}{$
\begin{split}
\mathcal{L}(X)  
& =  {\sum_{j = 1}^J \sum_{l = 1}^L{\mathbb{E}_{q(\thetav_j^{(1)}\given \xv_j)}}\left[ {\ln p\left({\bm{x}_{j}} \: | \:  \: \bm{\theta} _{j}^{(l)}\right)} \right]} -  {\sum_{j = 1}^J \sum_{l = 1}^L {{\mathbb{E}_{q(\thetav_j^{(>l)}\given \xv_j)}}\left[ {\ln \frac{{q\left(\bm{\theta} _{j}^{(l)} \: | \: \xv_{j}, \bm{\theta} _{j}^{(l+1)}  \right)}}{{p\left(\bm{\theta} _{j}^{(l)}\: |\: \: \bm{\theta} _{j}^{(l + 1)} \right) }}} \right]} } .
\end{split}$}
\end{equation} 
where the first term is the expected log-likelihood of the generative model, which ensures reconstruction performance, and the second term is the Killback–Leibler (KL) divergence that constrains the variational distribution $q(\thetav_j^{(l)})$ to be close to its prior $p(\thetav_j^{(l)})$.
The parameters in the Generalized GBN can be directly optimized by advanced gradient algorithms, like Adam \citep{kingma2014adam}.
The complete learning procedure of variational inference is summarized in Algorithm.~\ref{alg_aug_avi}.
\vspace{-1mm}
\subsection{Stable Training for the Generalized GBN}\vspace{-1mm}
\label{subsec_stable_traing}
Practical training of the Generalized GBN is highly challenging because of the objective's unbounded KL divergence \citep{razavi2019preventing, child2020very}.
In addition to applying the stable training techniques described in \citep{child2020very}, such as gradient skipping, we suggest three more technologies toward orienting the Generalized GBN.

% \textbf{Shape Parameter Clipping of Weibull distribution:} 
\textbf{Shape Parameter Clipping of Weibull distribution:}
As shown in Eq.~\eqref{eq_reparameterize}, when the sampled noise $\epsilonv$ is close to 1, e.g., $0.98$, and the Weibull shape parameter $k$ is less than $1e^{-3}$, the $x$ will be extremely huge, which could destabilize the training process. In practice, we constrain the shape parameter $\kv$ such that $1e^{-3}$ to avoid extreme value. 

\textbf{Weibull Distribution Reparameter:}
In our experiments, we reconstruct the inference network to stabilize the training.
Specifically, for the approximated Weibull posterior distribution, after inferring ${\small \kv_j^{(l)}}$, we let  ${\small \lambdav_j^{(l)} = \text{Softplus}\left( \fv_{\lambda}(\hv_j^{(l)}, \thetav_j^{(l+1)})\right)/\exp\left( 1+1/\kv_j \right)}$.
Specifically, for the latent variable $x\sim \mbox{Weibull}(k,\lambda )$, the expectation of latent variable $x$ is $ \lambda \mbox{exp}( 1+1/k)$. In this case, if $k$ is small, such as 0.001, the expectation of latent variable $x$ is $ \lambda \mbox{exp}( 1000)$, which is very layer for unstable training. 
To alleviate this challenge, we let the latent variable $x\sim \mbox{Weibull}(k, \lambda/\mbox{exp}( 1+1/k) )$. In this case, the expectation of latent variable $x$ is $ \lambda$, which is friendly with stable learning.

% \textbf{Learning Rate Decreasing:}
\textbf{Learning Rate Decreasing:}
After training a few epochs with the initialized learning rate, the training stage in our experiments will diverge. To achieve convergence in the training stage, the learning rate is reduced to 1/10 based on the initialization learning rate after some epochs.
\vspace{-2mm}
\section{Related Work}
\vspace{-2mm}
\textbf{Variational Autoencoder and its extension:} Gaussian VAE \citep{kingma2013auto, srivastava2017autoencoding}, have been used in different tasks such as image generation \citep{vahdat2020nvae}, graph generation \citep{kipf2016variational}, language model \citep{bowman2015generating}, time series forecasting \citep{krishnan2017structured}, out-of-distribution detection \citep{havtorn2021hierarchical}.
To improve the expressivity of the Gaussian VAE, there is a lot of effort put into developing deeper latent variable models
\citep{sonderby2016ladder, maaloe2019biva, dieng2019avoiding, vahdat2020nvae, child2020very, apostolopoulou2021deep}.
Apart from its expressive ability, the ability to disentangle data representation has also attracted wide attention \citep{higgins2016beta, burgess2018understanding}.
A simple method to enhance disentangling ability is to increase beta parameters \citep{higgins2016beta}, which may hurt the test reconstruction performance  \citep{kim2018disentangling}.
Unlike these works, the Generalized GBN learns to disentangle representation by its spares and non-negative latent variables.
Apart from the Gaussian VAE, other works have been proposed to model latent variables with Dirichlet distribution and sticking distribution \citep{joo2020dirichlet, nalisnick2016stick}.
However, these works are mainly a single-layer latent variable model, which does not directly extend to hierarchical structure.

\textbf{Gamma Belief Network and its variants:} As a full Bayesian generative model, the GBN \citep{zhou2015poisson, zhou2016augmentable} has greatly progressed in mining multi-layer text representation and extracting concepts.
With its attractive characters, there is a lot of effort to push it to adapt to different application scenarios.
Specifically,  \citep{guo2018deep} extend GBN to a deep dynamic system for temporal count data;  \citep{wang2019convolutional, wang2022generative} develop a convolutional GBN to capture word order information in text; and \cite{wang2020deep} model graph structure for document graph. 
Apart from the full Bayesian model that relies on Gibbs sampler for inference, \cite{zhang2018whai, zhang2020deep} build a Weibull deep autoencoder for GBN, which can utilize the neural network encoder. And \citep{duan2021sawtooth, li2022alleviating, duan2023bayesian} have tried to build a more effective and deeper GBN in the form of a variational autoencoder. 
However, all the above works use a linear decoder with $L_1$ norm and are limited to modeling complex, dense data, such as neural images.
\vspace{-2mm}
\section{Experiments}
\vspace{-2mm}
\subsection{Evaluating Expressiveness}\vspace{-1mm}
% \textbf{Datasets}
\textbf{Datasets:}
For binary images, we evaluate the models on two benchmark datasets: \textbf{MNIST} \citep{lecun1998gradient}, a
dataset of 28 × 28 images of handwritten digits, and \textbf{OMNIGLOT} \citep{lake2013one}, an alphabet
recognition dataset of 28 × 28 images. For convenience, we add two zero pixels to each border of the
training images. In both cases, the observations are dynamically binarized by being resampled from
the normalized real values using a Bernoulli distribution after each epoch, as suggested by \cite{lake2013one}, which prevents over-fitting. We use the standard splits of MNIST into 60,000 training
and 10,000 test examples, and of OMNIGLOT into 24,345 training and 8,070 test examples.
For natural images, we evaluate the models on two benchmark datasets: \textbf{CIFAR-10} \citep{krizhevsky2009learning}, a dataset of 32 × 32 natural images with ten classes, and \textbf{CELEBA} \citep{liu2015deep, larsen2016autoencoding}, a face dataset of 64 × 64.

% \vspace{1mm}
\textbf{Experiment Setting:}
% \textbf{Experiment Setting} 
For binary image datasets, we use a hierarchy of $L = 16$ variational layers, and the image decoder employs a Bernoulli distribution.
For neural images, we employ a hierarchy of $L=33$ variational layers for the CIFAR-10 dataset and $L=42$ variational layers for the CELEBA dataset.
Meanwhile, we use a mixture of discretized logistic distributions \citep{salimans2017pixelcnn++} for the data distribution.
It should be noted that Generalized GBN's code is built on the VDVAE codebase\footnote{\url{https://github.com/openai/vdvae}}\citep{child2020very}, with minor modifications made to the variational Weibull posterior and loss functions.
% The more detailed hyperparameters, such as learning rate, can be found in Appendix.~\ref{app_detailed_hyperparameter}. 
All experiments are performed on
workstation equipped with a CPU i7-10700 and accelerated
by four GPU NVIDIA RTX 3090 with 24GB VRAM. 

% \textbf{Benchmark Models} 
\textbf{Benchmark Models:}
We conducted a comparison between our model and the state-of-the-art variational autoencoder, which is comprised of AttnVAE \citep{apostolopoulou2021deep}, VDVAE\citep{child2020very}, and NVAE\citep{vahdat2020nvae}, as proposed recently.
Additionally, the DirVAE\citep{joo2020dirichlet} and SBVAE\citep{nalisnick2016stick}, which model latent variables with the Stick-breaking distribution and Dirichlet distribution, respectively, also serve as baseline models. 
\begin{table*}[t]
\caption{\small Comparison against the state-of-the-art likelihood-based generative models. The performance
is measured in bits/dimension (bpd) for  the CIFAR-10 and CELEBA datasets, but MNIST and OMNIGLOT in which negative log-likelihood
in nats is reported 
(Lower is better in all cases.). The marginal loglikelihood of the MNIST and OMNIGLOT datasets is estimated
with 1000 important samples.  
( The marginal loglikelihood of Attention VAE is estimated with 100 important samples on CIFAR-10 dataset.) 
\label{tab:fid_is}}
\begin{adjustbox}{width=1\textwidth,
center}
\begin{tabular}{clcccc}
 \toprule
 Model type & Model & MNIST & OMNIGLOT & CIFAR-10  & CELEBA \\[1pt] %
 \midrule
 \textbf{Gamma VAE Models} & \textbf{Generalized GBN} & 77.86 & \textbf{86.04} & 2.84 &\textbf{1.95} \\[1pt]
 \midrule
 \multirow{6}{*}{\tabincell{c}{VAE Models with \\ an Unconditional Decoder }} 
 &Attention VAE~\citep{apostolopoulou2021deep} & \textbf{77.63} & 89.50 &\textbf{2.79} & - \\[1pt]
 &VDVAE~\citep{child2020very} & 78.07 & 86.93 & 2.87& 2.00\\[1pt] %
 & NVAE~\citep{vahdat2020nvae} & 78.19 & 90.18 & 2.91&2.03 \\[1pt]
 &BIVA \citep{maaloe2019biva} & 78.41& 93.54 & 3.08 & 2.48 \\[1pt]
 &DVAE++ ~\citep{vahdat2018dvae++} & 78.49 & 97.43 & 3.38 & \\[1pt]
 & Conv DRAW \cite{gregor2016towards} & - & 91.00 & 3.58 & - \\[1pt]
 \midrule
 \multirow{6}{*}{\tabincell{c}{Flow Models without any \\[1pt] Autoregressive Components }} 
 & VFlow \citep{chen2020vflow} & - & -&2.98 &- \\[1pt] %
 & ANF \citep{huang2020augmented}& -& -& 3.05&  \\[1pt]
 &  Flow++  \citep{ho2019flow++}& - &- & 3.08 & -\\[1pt]
 &  Residual flow \citep{chen2019residual}& - & - &3.28 & -  \\[1pt]
 & Real NVP \citep{dinh2016density} &-&-&3.49&3.02 \\[1pt] 
 \midrule 
 \multirow{6}{*}{\tabincell{c}{ VAE and Flow Models with \\[1pt] Autoregressive Components}} &  $\delta$-VAE \citep{razavi2019preventing}& -& -&2.83&  \\[1pt]
 & PixelVAE++ \citep{sadeghi2019pixelvae++}&78.00 &- &2.90\\[1pt]
 & VampPrior \citep{tomczak2018vae} &78.45 & - &- &- \\[1pt]
 & MAE \citep{ma2019mae}& 77.98& - &2.95&-\\[1pt]
 & Lossy VAE \citep{chen2016variational} &78.53& 89.83&2.95&-\\[1pt]
 & MaCow \citep{ma2019macow}&-&-&3.16&  \\[1pt]
\midrule
 \multirow{5}{*}{Autoregressive Models} & PixelCNN++ ~\citep{salimans2017pixelcnn++} & -& -&2.92& - \\[1pt] %
 & PixelRNN ~\citep{van2016pixel} & -& -&3.00&- \\[1pt] %
 & Image Transformer \citep{parmar2018image} & -& -&2.90& - \\[1pt]
 & PixelSNAIL \citep{chen2018pixelsnail} & - &- &2.85&-\\[1pt]
 & Gated PixelCNN \citep{van2016conditional} & - & - &3.03 &-\\[1pt]
 \midrule
 \multirow{2}{*}{Non-Gaussian VAE Models}& DirVAE \citep{joo2020dirichlet} & 84.76 &  95.82 & - & -  \\[1pt]
 &SBVAE\citep{nalisnick2016stick} & 99.27 & 128.82  & - & - \\[1pt]   
 \bottomrule
\end{tabular}
\vspace{-5mm}
\end{adjustbox}\vspace{-5mm}
\end{table*}
\begin{figure}[h]
  \begin{minipage}{0.5\textwidth}
    \centering
\subfigure[Reconstruction]
{
\includegraphics[width=0.4\linewidth]{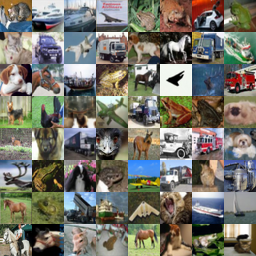}
\label{fig_cifar10_recon}
}
\quad 
\subfigure[Sample]{
\includegraphics[width=0.4\linewidth]{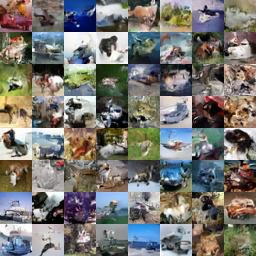}
\label{fig_cifar10_sample}
} \vspace{-2mm}
\caption{Reconstruction and unconditional generation samples of
the Generalized GBN.
} \label{fig_cifar10_vis}
  \end{minipage}\quad \quad
  \begin{minipage}{0.4\textwidth}
    \centering
  \centering
  \renewcommand{\arraystretch}{1.2}
  \begin{adjustbox}{width=0.8\textwidth,
center}
  \begin{tabular}{llc}
    \toprule
    \multicolumn{2}{l}{Model} &FID$\downarrow$\\
    \midrule
    PixelCNN  &&65.9\\
    Glow  &&48.9\\
    NVAE  &&51.7\\
    VDVAE &&40.1\\
    \midrule
    \textbf{Generalized GBN} &&\textbf{38.4}\\
    \bottomrule
    \end{tabular}
    \end{adjustbox}
    \caption{FID scores for unconditional generation on CIFAR-10}
    \label{fid_score}
  \end{minipage}  
\vspace{-3mm}
\end{figure}
% \textbf{Experiment Results}

\textbf{Experiment Results:}
Table.~\ref{tab:fid_is} reports the estimated marginal likelihood of our model along with the performance
achieved by state-of-the-art models.
We observe that in both datasets, the Generalized GBN
consistently improves performance to different degrees than the base model VDVAE.
These results confirm our motivation for the Generalized GBN to enhance its expressivity with a more powerful generative model.
Further, the comparable performance of the Generalized GBN and hierarchical Gaussian VAE may be attributed to the fact that dense latent variables and sparse non-negative latent variables are integral parts of neural networks, as illustrated in Fig.~\ref{motivation}.
Correspondingly, Dirichlet VAE's performance is relatively poor, partly due to shallow's generative model, but more importantly, the $L_1$ norm on its latent variables would limit the expression ability \citep{shen2023study}.
Thanks to its powerful attention mechanism, the Attention VAE performs superiorly on Mnist datasets.
Additionally, it may be unfair to compare our results with Attention VAE on CIFAR-10 data, as the latter employs 100 important samples whereas ours uses only one. More results can be found in \ref{ppl_comare_section}.

\textbf{Unconditional Image Generation Results:} 
In order to evaluate the unconditional image generation ability of the proposed model, we further take the unconditional image generation experiments on the widely used CIFAR-10 dataset \citep{krizhevsky2009learning}.
The FID scores for unconditional generation are shown in Fig.~\ref{fid_score}, and corresponding unconditional generate samples are shown in Fig.~\ref{fig_cifar10_vis}. 
The results show that the generalized GBN can generate meaningful samples and achieve better performance compared with strong baselines such as NVAE and VDVAE on the unconditional image generation task.
\subsection{Evaluating Disentangled Representations}
\textbf{Dataset:}
We compare the Generalized GBN to various baseline models on the following data sets:
% with (i) known generative factors: 
1) \textbf{2D Shapes} \citep{matthey2017dsprites}: 737,280 binary 64 × 64 images of 2D shapes
with ground truth factors [number of values]: shape[3],
scale[6], orientation[40], $x$-position[32], $y$-position[32].
2) \textbf{3D Shapes} \citep{3dshapes18}: 480,000 RGB
64 × 64 × 3 images of 3D shapes with ground truth factors:
shape[4], scale[8], orientation[15], floor colour[10], wall
colour[10], object colour[10] 
$ii)$ unknown generative factors:  \textbf{3D Cars} \citep{reed2015deep}: 286,560 RGB 64 × 64 × 3 images of car CAD models.

% \textbf{Evaluation Matric} 
\textbf{Evaluation Matric:}
The disentanglement in BetaVAE metric \citep{higgins2016beta} is measured as the accuracy of a linear classifier that predicts the index of a fixed factor of variation.
To address several issues in the BetaVAE metric, \cite{kim2018disentangling} develop the FactorVAE metric by using a majority vote classifier
on a different feature vector 
 which accounts for a corner
case in the BetaVAE metric.
Differently, the
Modularity \citep{ridgeway2018learning} measures disentanglement if each
dimension of the learned
representation depends on at most a factor of variation
using their mutual information. The Disentanglement metric
of \citep{eastwood2018framework} ( DCI Disentanglement) computes the entropy of the distribution obtained by normalizing the importance of each dimension of the learned representation for predicting the value of
a factor of variation.
The SAP score \citep{kumar2017variational} is
the average difference in the prediction error of the two most
predictive latent dimensions for each factor.
\begin{table*}[t]
\caption{\small Comparison against the state-of-the-art disentangled representation learning generative models. 
\label{tab:disangled_result}}
\begin{adjustbox}{width=1\textwidth,
center}
\begin{tabular}{clccccccc}
 \toprule
 Dataset & Model & $\betav$-VAE  & FactorVAE & DCI-D & DCI-I& DCI-C & Modularity& SAP \\ %
 \midrule
 \multirow{6}{*}{\tabincell{c}{2D Shapes}} 
 &VAE~\citep{kingma2013auto} & 0.778 & 0.617 & 0.103 & 0.393& 0.100 & 0.793 &0.032 \\
 &$\betav$-VAE~\citep{higgins2016beta} & 0.817 & 0.597 & 0.238 & 0.501 & 0.266 & 0.820 & 0.065 \\[1pt] %
& Factor VAE ~\citep{kim2018disentangling} & 0.871 & 0.746 & 0.238 & 0.499 & 0.205  &0.775 &0.074 \\[1pt]
 &BetaTCVAE ~\citep{chen2018isolating} & 0.884 & 0.759 &0.304 & 0.567 & 0.304 &0.877 & 0.069 \\[1pt]  
 & DIP-VAE ~\citep{kumar2017variational} & 0.855 & 0.701 & 0.157 & 0.402 & 0.161 & 0.872 & 0.043\\[1pt]
& \textbf{Generalized GBN} & 0.853
& 0.626 & 0.187 & 0.433 & 0.271 & \textbf{0.888} & 0.058\\[1pt]
 \midrule
 \multirow{6}{*}{\tabincell{c}{3D Shapes }} 
 &VAE~\citep{kingma2013auto} & 0.732 & 0.550 & 0.156 & 0.649 & 0.135 & 0.847 & 0.012 \\[1pt]
 &$\betav$-VAE~\citep{higgins2016beta} & 0.999 & 0.802 & 0.539 & 0.920 & 0.455 & 0.933 & 0.092 \\[1pt] %
& Factor VAE ~\citep{kim2018disentangling} & 0.999 & 0.844 & 0.642 & 0.910 & 0.530 & 0.977 & 0.106\\[1pt]
 &BetaTCVAE ~\citep{chen2018isolating} & 1.000 & 0.974 & 0.876  & 0.992 & 0.761 & 0.962 & 0.097 \\[1pt] 
 & DIP-VAE ~\citep{kumar2017variational} & 0.976 & 0.921 & 0.687 & 0.900 & 0.594 & 0.932 & 0.089\\[1pt] 
 & \textbf{Generalized GBN} & 0.912 & 0.841 & 0.621 & 0.906 & 0.514 & 0.942 & 0.048\\[1pt]
 \midrule 
 \multirow{6}{*}{\tabincell{c}{ 3D Cars}} & VAE~\citep{kingma2013auto} & 0.999 & 0.753 & 0.110 & 0.702 & 0.070 & 0.836 & 0.027 \\[1pt]
 &$\betav$-VAE~\citep{higgins2016beta} & 1.000 & 0.923 & 0.316 & 0.559 & 0.224 & 0.930 & 0.005\\[1pt] %
& Factor VAE ~\citep{kim2018disentangling} & 1.000 & 0.907 & 0.144 &0.682 & 0.142 & 0.889 & 0.009\\[1pt]
 &BetaTCVAE ~\citep{chen2018isolating} & 1.000 & 0.929 & 0.318 & 0.817 & 0.246 & 0.931 & 0.014\\[1pt] 
 & DIP-VAE ~\citep{kumar2017variational} & 1.000 & 0.873 & 0.261 & 0.704 & 0.142 & 0.837 & 0.014\\[1pt]
 & \textbf{Generalized GBN}& \textbf{1.000} & \textbf{0.946} & \textbf{0.323} & 0.813 & 0.218 & 0.897 & 0.023\\[1pt]
 \bottomrule
\end{tabular}\vspace{-4mm}
\end{adjustbox}%
\vspace{-4mm}
\end{table*}

\textbf{Baseline Models:}
The baseline models consist of the base VAE \citep{kingma2013auto} and
the methods by which the training loss is augmented with a regularizer, including the $\betav$-VAE \citep{higgins2016beta},
introduce a hyperparameter in front of the KL regularizer of vanilla VAEs to constrain the capacity of the
VAE bottleneck.
The FactorVAE \citep{kim2018disentangling} penalize the total correlation \citep{watanabe1960information} with adversarial training \citep{nguyen2010estimating, sugiyama2012density}; 
and the
Beta-TCVAE \citep{chen2018isolating}  with a tractable but
biased Monte-Carlo estimator.
The DIP-VAE \citep{kumar2017variational} penalize the
mismatch between the aggregated posterior and a factorized
prior. 
And all the experiments are taken with the open codebase \citep{locatello2019challenging}\footnote{\url{https://github.com/google-research/disentanglement_lib/tree/master}}.
The generalized GBN, based on the same codebase, has a slight difference in the Weibull variational posterior and loss functions.
% The more detailed hyperparameters can be found in Appendix.~\ref{app_detailed_hyperparameter}. 

\textbf{Quantitative Results:}
The experiment results on different evaluation matric of disentanglement are listed in Table.~\ref{tab:disangled_result}.
First of all, the results indicate that the Generalized GBN outperforms the basic Gaussian VAE, which retains the original VAE loss without any regularization.
Secondly, the Generalized GBN achieves comparable performance compared with state-of-the-art disentangled models.
It's noted that the improved disentangled models based on Gaussian VAE often need different regularizations, which may negatively impact test reconstruction ability \citep{kim2018disentangling}.
However, the Generalized GBN relies directly on the gamma latent variable's sparseness and does not apply more regularization.
On the other hand, these regularizations could also potentially improve Generalized GBN's performance.
Furthermore, the Generalized GBN  has the potential to learn hierarchical disentangled latent representations due to its hierarchical sparse gamma latent variables \citep{zhou2015poisson,ross2021benchmarks}.
% Besides, different from Gaussian VAE, which relies on the constrained information bottlenecks \citep{burgess2018understanding} to learn disentangled representation,  the Generalized GBN relies on gamma latent variables' sparseness and can potentially model hierarchical disentangled latent representations \citep{zhou2015poisson,ross2021benchmarks} for future works.

\begin{figure*}[t]
\centering
\subfigure[Generalized GBN]
{
\includegraphics[width=0.42\linewidth]{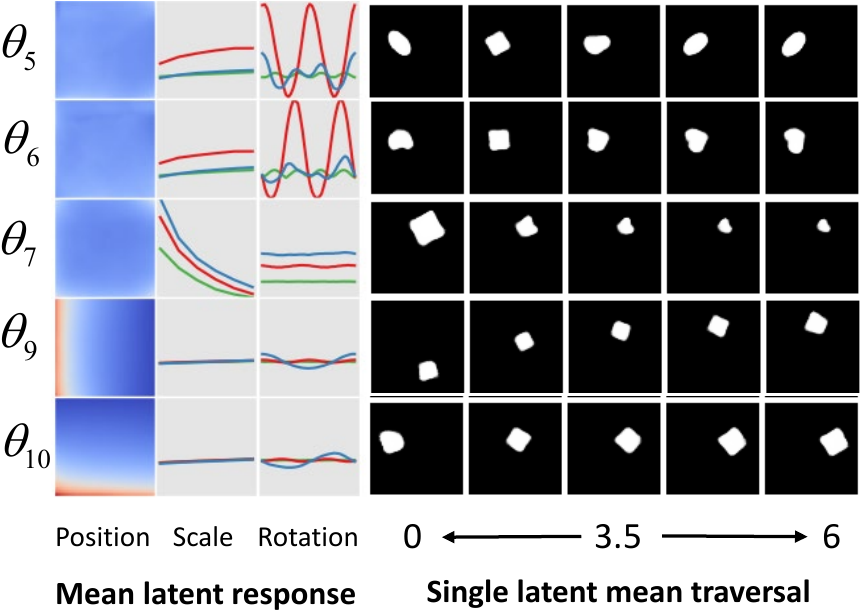}
\label{fig_embedding_vis_1}

}
\quad \quad
\subfigure[Gaussian VAE]{
\includegraphics[width=0.405\linewidth]{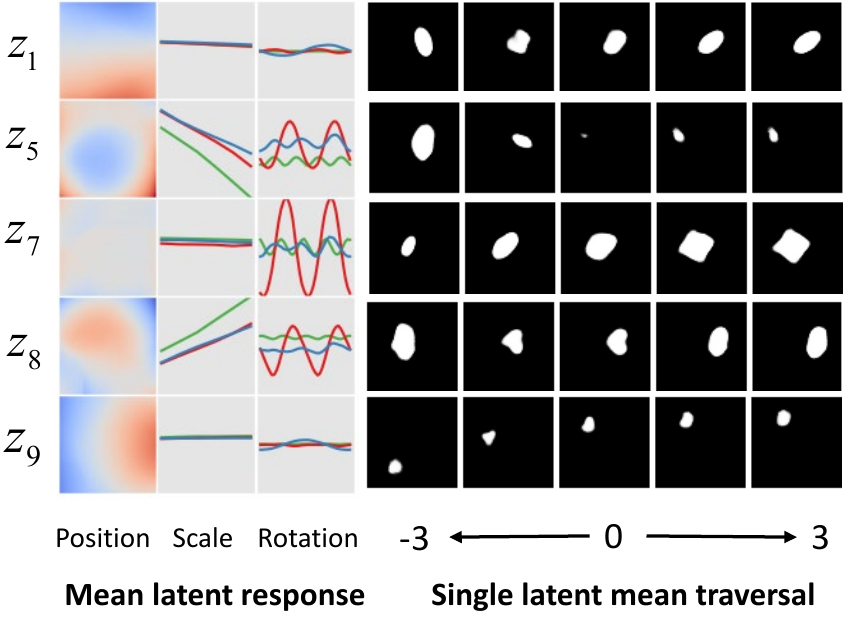}
\label{fig_embedding_vis_2}
} \vspace{-2mm}
\caption{\textit{\ref{fig_embedding_vis_1}}: Representations learned by
the Generalized GBN.
Each row represents a latent $\theta_i$.
Column 1 (position) shows the mean activation (red represents high values) of
each latent $\theta_i$ as a function of all 32x32
locations averaged across objects, rotations and scales. Columns 2 and 3 show the mean activation of each unit $\theta_i$ as a function of scale (respectively rotation), averaged across rotations and positions (respectively scales and positions). 
Square
is red, oval is green and heart is blue.
Columns 4-8 (second group) show reconstructions resulting from the traversal
of each latent $\theta_i$ over 0 to 6 while keeping the remaining 9/10
latent units fixed to the values obtained
by running inference on an image from
the dataset. \textit{\ref{fig_embedding_vis_2}}: Similar analysis for Gaussian VAE.
} \label{fig_disentangled}
\vspace{-3mm}
\end{figure*} 
% \textbf{Qualitative Result}
\textbf{Qualitative Result:}
The qualitative results of the Gaussian VAE and the Generalized GBN disentangled representations are depicted in Fig.~\ref{fig_disentangled}.
Firstly, as the top three columns indicate, each dimension of latent variables in Generalized GBN has distinct semantics.
In particular, position $y$ and $x$ are represented by $\theta_{9}$, $\theta_{10}$, and scale is represented by $\theta_{7}$, respectively.
However, the semantics of latent variables in Gaussian VAE are coupled, such as $z_{8}$, which simultaneously represents position, scale, and rotation.
Secondly, columns 4-8 of Generalized GBN indicate that $\theta_7$, $\theta_9$, and $\theta_{10}$ can control the data's scale and position, respectively. 
However, columns 4–8 of Gaussian VAE
show the couped semantics.
Consequently, it is evident that the Generalized GBN outperforms Gaussian VAE in the disentangled ability, whereby each dimension latent variable influences distinct semantic structures.
The qualitative result can also confirm the results of the quantitative results that are displayed in Table.~\ref{tab:disangled_result}
\vspace{-2mm}.
\section{Conclusion}
\vspace{-2mm}
In this paper, we introduce the Generalized GBN, which extends the capabilities of the GBN by incorporating a more expressive non-linear generative model. To enable effective inference, we develop a Weibull variational upward-downward inference network to approximate the posterior distribution of latent variables. To assess the model's expressivity, we conduct extensive experiments on benchmark datasets, utilizing test likelihood as a metric. Our experimental results demonstrate that removing the linear decoder limitation of GBN can significantly enhance the model's modeling capabilities, achieving comparable performance with state-of-the-art Gaussian VAEs. 
Furthermore, in disentangled representation learning experiments, the Generalized GBN exhibits strong performance compared to various baseline models. 
This outstanding performance can be attributed to the sparse and non-negative properties of the gamma latent variables.
% For future works, according to recent studies \citep{dai2021knowledge, geva2020transformer}, non-negative vectors provide better identification of knowledge neurons in transformers \citep{vaswani2017attention}, which opens up greater potential for sparse and non-negative gamma latent variables.

% \begin{table}[h]
%   \caption{FID scores for unconditional generation on CIFAR-10}
%   \label{fid_score}
%   \centering
%   \begin{tabular}{llc}
%     \toprule
%     \multicolumn{2}{l}{Model} &FID$\downarrow$\\
%     \midrule
%     PixelCNN  &&65.93\\
%     Glow  &&48.9\\
%     NVAE \citep{vahdat2020nvae}  &&51.67\\
%     VDVAE \citep{child2020very} &&40.1 \\
%     \midrule
%     \textbf{Generalized GBN} &&\textbf{38.4}\\
%     \bottomrule
%   \end{tabular}
% \end{table}

% \begin{figure*}[ht]
% \centering
% \subfigure[Reconstruction]
% {
% \includegraphics[width=0.22\linewidth]{fig/cifar10_recon.png}
% \label{fig_cifar10_recon}

% }
% \quad \quad
% \subfigure[Sample]{
% \includegraphics[width=0.22\linewidth]{fig/cifar10_sample.png}
% \label{fig_cifar10_sample}
% } 
% \caption{Representations learned by
% the Generalized GBN.
% } \label{fig_cifar10_vis}
% \vspace{-2mm}
% \end{figure*}

\bibliography{example_paper.bib}
\bibliographystyle{unsrtnat}

%%%%%%%%%%%%%%%%%%%%%%%%%%%%%%%%%%%%%%%%%%%%%%%%%%%%%%%%%%%%
\newpage
\appendix
\section{Perplexity comparision}\label{ppl_comare_section}
We conducted document modeling experiments on three popular text datasets to evaluate the improvement achieved by replacing a linear decoder with a non-linear neural network decoder. Table~\ref{ppl_compare} shows that GGBN outperforms other baseline models, indicating that GGBN is an effective method for modeling text data, such as Bag-of-words. It should be emphasized that GGBN can also generalize to natural images and other complex data, while traditional GBN-like baselines are limited to count data only.
\begin{table}[th!]
    \centering
    \caption{Perplexity (the lower, the better) results on three popular text datasets between GGBN and popular GBN-like model. The experimental settings and some baseline results follow the work of \citet{duan2023bayesian}. }
    \label{ppl_compare}
    \begin{tabular}{lcccc}
    \toprule
         Methods&Depth &20NG &RCV1 &R8   \\
         LDA &1 & 735 &942 &966\\
         ProdLDA &1 & 784 &951 &561\\
         ETM &1 & 742 &921 &985\\
         GBN &5 & 678 &877 &657\\
         WHAI &5 & 726 &906 &773\\
         SawETM &5 & 685 &873 &530\\
         dc-ETM &5 & 647 &801 &420\\
         ProGBN-x &5 & 653 &798 &436\\
         ProGBN-kg &5 & 620 &753 &411\\
         ProGBN-wv &5 & 614 &735 &408\\
         \textbf{Generlized GBN} &5 & \textbf{589} &\textbf{685} &\textbf{385}\\
    \bottomrule
    \end{tabular}

\end{table}

\section{Training Algorithim}

\begin{algorithm}[h]
   \caption{Upward-Downward Variational Inference}
   \label{alg_aug_avi}
   \begin{algorithmic}
   \STATE \textbf{Input:} Observed data $X=\{x_n\}_{j}^{J}$.
   \STATE \textbf{Output:} Global parameters of the Generlized GBN $\Wv^{(l:L)}$.
   \STATE Set mini-batch size $m$ and the number of layer $L$
   \STATE Initialize the parameters $\Wv^{(l:L)}$;
   \FOR{$\text{iter = 1,2,} \cdot \cdot \cdot $} 
   \STATE 1. Randomly select a mini-batch of $m$ documents to form a subset ${X} = {\left\{ {{\xv_i}} \right\}_{1,m}}$;\vspace{0.5mm}
   \STATE 2. Infer the variational posterior for gamma latent variable $\{ \thetav _i^{(l)}\}_{i=1,l=1}^{m,L-1}$ with the inference network via  Eq. (\ref{Eq_inference}) ; \vspace{0.5mm}
   \STATE 3. Drawn random noise $\left\{ {{\bm{\varepsilon} _i^l}} \right\}_{i = 1,l = 1}^{m,L}$ from a uniform distribution;\vspace{0.5mm}
   \STATE 4. Sample hierarchical latent variables $\{ \thetav _i^{(l)}\}_{i=1,l=1}^{m,L-1}$ from Weibull distribution with $\left\{ {{\bm{\varepsilon} _i^l}} \right\}_{i = 1,l = 1}^{m,L}$ via Eq. (\ref{eq_reparameterize});
   \vspace{0.5mm}
    \STATE 5. Calculate  $\nabla {}_{\Wv^{(l:L)}}\mathcal{L}\left( {\Wv^{(l:L)};{X}; \left\{ {{\bm{\varepsilon} _i^l}} \right\}_{i = 1,l = 1}^{m,L}} \right)$ according to Eq.~\eqref{aug_ELBO}, and update encoder parameters and decoder parameters $\small \Wv^{(l:L)}$ jointly ;
   \ENDFOR
\end{algorithmic}
\end{algorithm}  % \linespread{1}

\end{document}